\documentclass[letterpaper, 10 pt, conference]{ieeeconf}  

\IEEEoverridecommandlockouts                              
\usepackage{graphics} 
\usepackage{multirow}
\usepackage{color}
\usepackage{graphicx}
\usepackage{amssymb}
\usepackage{latexsym}

\usepackage{authblk}
\usepackage[utf8]{inputenc}
\usepackage[english]{babel}

\usepackage{blindtext}

\title{\LARGE \bf
Widening siamese architectures for stereo matching
}

\author{Patrick Brandao$^{}$, Evangelos Mazomenos$^{}$, Danail Stoyanov$^{}$
\thanks{Corresponding author: patrick.brandao.15@ucl.ac.uk}
}
\affil[]{Centre for Medical Image Computing and Department of Computer Science, University College London, UK}

\begin{document}

\maketitle
\thispagestyle{plain}
\pagestyle{plain}

\begin{abstract}

Computational stereo is one of the classical problems in computer vision. Numerous algorithms and solutions have been reported in recent years focusing on developing methods for computing similarity, aggregating it to obtain spatial support and finally optimizing an energy function to find the final disparity. In this paper, we focus on the feature extraction component of stereo matching architecture and we show standard CNNs operation can be used to improve the quality of the features used to find point correspondences. Furthermore, we propose a simple space aggregation that hugely simplifies the correlation learning problem. Our results on benchmark data are compelling and show promising potential even without refining the solution.

\end{abstract}

\section{Introduction}
\label{sec1}

Computational stereo is one of the classical problems in computer vision systems whereby two cameras placed at different viewpoints can be used to extract 3D information by analyzing the relative position of the objects in the two perspectives of the scene. Finding relative displacements between image pairs from stereo cameras is usually called stereo matching \cite{scharstein2002taxonomy,brown2003advances}. By using the fundamental constrains in the two-view geometry of two perspective cameras, it is possible to reduce the stereo matching problem to a 1D search space in horizontally rectified images. Despite the reduced search space, accurately finding stereo correspondences in real world images is still very challenging because occlusions, reflective surfaces, repetitive patterns, textureless or low detail regions that can affect the similarity metric and underpins the search.  

Recently, since the first winning entry in the ImageNet Large Scale Visual Recognition Challenge, deep learning has been at the forefront of most computer vision breakthroughs \cite{russakovsky2015imagenet}. Convolution neural networks (CNNs) are able to learn very complex non-linear representations from raw visual data, creating effective and versatile models for complex problems. CNNs are now widely used across different vision problems and also in a vast range of applications, such as robotics and medical endoscopic imaging. Deep learning models have also recently been applied to stereo matching and are now among the most accurate methods reported on the public common evaluation datasets \cite{Geiger2012CVPR,Menze2015CVPR,scharstein2014high}.

\begin{figure*}[t]
    \centering
    \includegraphics[width=\textwidth]{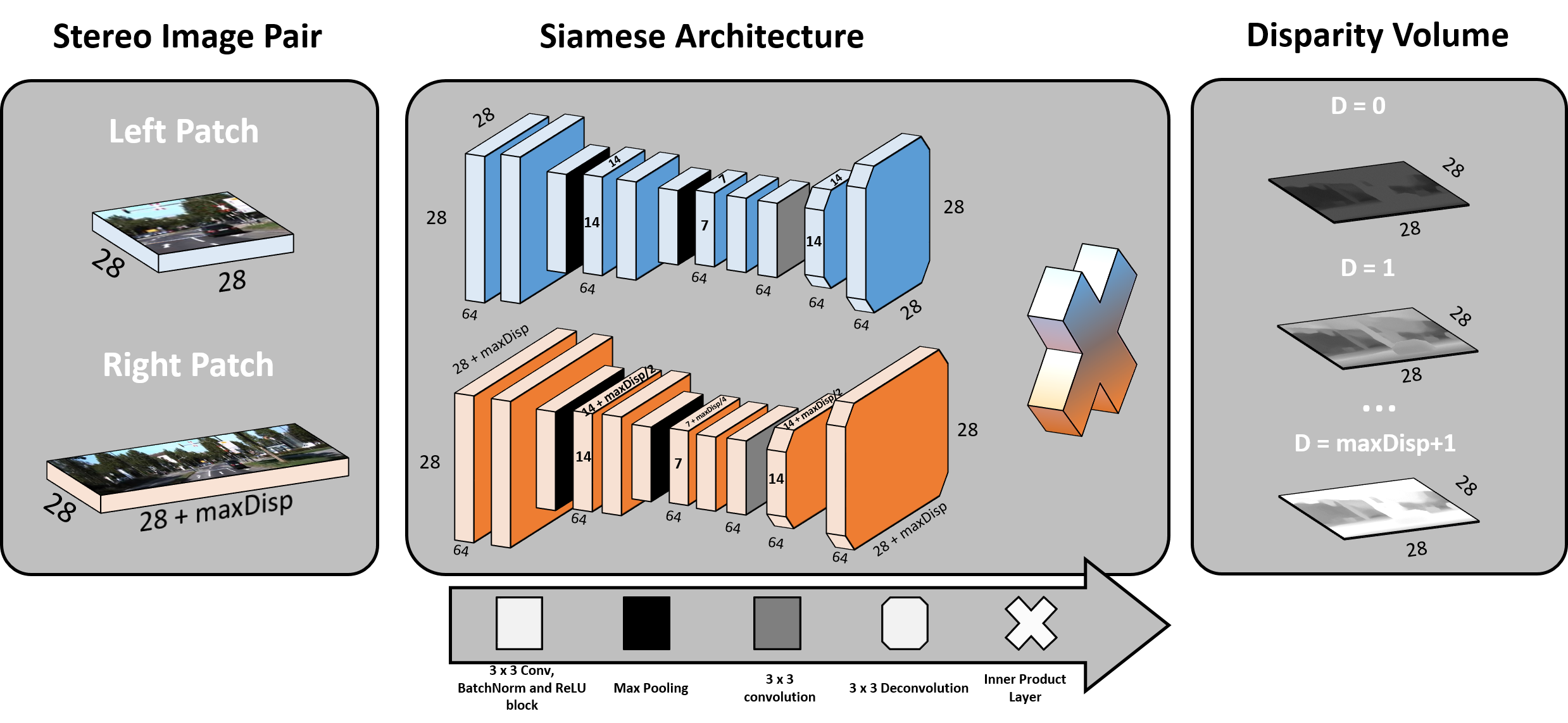}
    \caption{Representation of our 7 layered stereo matching CNN. Patches extracted from the left and right stereo images are processed in the blue and orange branches, respectively. During training, the width of the right patch depends of the max disparity ($D$) considered. After feature extraction with the siamease architecture, the features are aggregated according to their relative displacement. The correlation between features for each disparity is computed by a simple two layer correlation architecture. The final disparity volume represents a correlation value of each possible integer disparity between zero and $D$ for every left patch pixel. }
    \label{fig:maingraph}
\end{figure*}

One of first successful uses of deep learning for stereo matching treats the problem as a binary classification \cite{zbontar2016stereo}, where different CNNs are trained to recognize two input patches centered around corresponding pixels. However, because each pixel is processed individually, and no spatial  constrains are imposed in the decision, the resulting disparity map can be quite noisy. To mitigate noise, extensive post processing steps are used to smooth the result using hand-crafted regularization functions. Several improvements have been reported since then, usually by stacking extra convolution layers after the feature extraction, allowing the CNNs to learn their own spatial regularization. The current top stereo method ranked on the KITTI benchmark dataset also focuses on context and consistency by using a very deep end-to-end learned architecture with 3-D convolutions that is able to infer disparity maps capable of beating any hand-crafted regularized method \cite{kendall2017end}. 

While spatial consistency is essential for good stereo matching, there has been limited focus on the quality of the high-level representation learned to match corresponding points. Several methods proposed different architectures, correlation operations or regularization approaches but the majority of CNN stereo methods start with a relatively shallow siamese architecture that acts as a feature extractor for the stereo image pair. In this work, we take a step back from deep complex CNN architectures and focus on the type of features that are used to find correspondences. We propose the use of pooling and deconvolution operations in the siamease architecture that allows the extraction of features with a wider receptive field around the target pixels. The intuition is that, a wider context view allows the feature extraction of more visual cues, allowing better point correspondence. Furthermore, we propose a simple feature space transformation that significantly simplifies the learning problem, allowing the CNNs to learn end-to-end correlation with a very shallow architecture. Our main objective is to show that improvements can be achieved simply by enhancing the way stereo features are extracted and aggregated. Because siamese architectures are part of most matching CNNs available, this work can easily be combined with more complex approaches (hand-crafted or deep learned) present in the literature.

\section{Related work}
\label{sec:relatedwork}

A huge range of approaches has been proposed to solve the stereo matching problem in the last decades. For the sake of brevity, we will focus on the work that exploits deep learning as a viable way to find point correspondences in image pairs \cite{scharstein2002taxonomy,brown2003advances}.

The introduction of large scale, high resolution datasets, such as KITTI \cite{Geiger2012CVPR,Menze2015CVPR} and Middlebury  \cite{scharstein2014high}, has opened the opportunity for the use of learning approaches in stereo matching. As stated before, \cite{zbontar2016stereo} used a siamese CNN to binary classify matching or non-matching pairs of points. The method required an extensive post processing step, where edge and texture information were used as smoothness constrains. 

More recently, \cite{luo2016efficient} expanded on Zbontar's work and proposed a way to obtain disparity values for all possible displacements without manually pairing patch candidates. In other words, a wider image is passed though one of the branches of the siamese architecture and the computed features are correlated with the ones extracted from the target patch. This allows the computation of matching costs for all disparities with one-pass of the CNN. This work also shows that the inner product is a fast and effective way to compute feature correlation. Again, because inference for each pixel is made independently, hand-crafted feature regularization is used to smooth the results.

Currently, the top performing stereo methods in the KITTI datasets \cite{Geiger2012CVPR,Menze2015CVPR} focus on end-to-end network learning with spatial regularization and do not use any type of hand-crafted post processing. \cite{shaked2016improved} employ a second network that is trained to smooth the matching cost obtained by a deep residual architecture. Kendall \textit{et al.} use 37 layered network with multi-scale 3D convolutions to learn how to match a block of concatenated features from both images. \cite{pang2017cascade}  tackled the matching problem in two stages: first, a tweaked version of DispNet is used to estimate disparities with more detail and then a second network is used to rectify the results of the first stage. \cite{DBLP:journals/corr/KnobelreiterRSP16} also achieved excellent performance by combining CNNs and conditional random fields into a hybrid model for stereo estimation. Despite the huge difference in architectures and training methodology, all these  methods start roughly the same way, with a siamease architecture that acts as a feature descriptor for the stereo image pair. Most recent work chooses to focus on the spatial regularization rather than the feature extraction step. We argue that significant improvements can be achieved by simply increasing the amount of context that is extracted by the siamease architecture. 

The work presented here is most similar to the one developed by Luo et al. \cite{luo2016efficient} but with two major contributions. First, we show that the loss of the detail from pooling operations can be compensated with deconvolution operations if these are applied in the feature space, before computing correlation. This allows to hugely increase the global receptive field of the feature extractors, resulting in a more robust matching even before spatial regularization. Second, we show that a simple feature aggregation can be used to simplify the learning problem, resulting in effective, more easily learned, data driven correlation metric. To reiterate, because we are just proposing to improve the feature extraction step, our aim is not to beat the current state-of-the-art for full stereo matching pipelines. Our contribution provides a very effective and fast stereo matching network that can easily be further improved by plugging it to most current CNN stereo matching models. 

\section{Methodology}
\label{sec:methods}

Let us consider a stereo image pair $I^L,I^R$ with size $W\times{H}$. A typical stereo algorithm computes a cost volume $C$ such that:
 
 \begin{equation}
C(x,y,d)=\sum_{d}^{D}\sum_{x}^{W}\sum_{y}^{Y}   s(I^L(x,y),I^R(x,y-d)) .  \label{eq1}
\end{equation}

where $D$ is the maximum disparity under consideration and the function $s$ returns a similarity score between the two pixels that are indexed in the horizontal and vertical directions by $x$ and $y$. Typically, $s$ is a similarity function between handcrafted representations of small patches around the pixels \cite{brown2003advances}. Alternatively, CNNs can be use to learn complex, high dimensional feature extractors that allow a more robust patch comparison \cite{zbontar2016stereo}.         

Some of the most accurate stereo algorithms proposed in recent years employ CNNs to score the patch similarity measure \cite{park2016look,zbontar2016stereo,luo2016efficient,shaked2016improved,kendall2017end}. Even though these methods proceed with different approaches, every model starts with a siamese architecture that processes the left and the right images. While subsequent layers may allow more complex correlation inference or spatial regularization of the cost volume, the matching is still in essence based on the features extracted by the siamese branches. As a consequence, the architecture of the siamese CNN plays a crucial role in the quality of the stereo matching, much like the role of a traditional low level vision similarity metric. We therefore focus on enhancing the underlying siamese network in order to improve performance.

\subsection{Siamese network architecture}

We construct our network by layering sequential blocks of 2D convolutions, batch normalization and a rectifier linear unit (ReLU). Just like most architectures, we use layers with 64 neurons of $3\times{3}$  convolutions and the parameters between branches are shared. The last layers are added without batch normalization and ReLU operations. 

Generally speaking, wider patches allow the extraction of more visual cues and help more accurate matching, especially in textureless regions or areas of aperture problems. The area around the target pixel that is considered in the matching process depends on the global receptive field of the CNN architecture. If we denote the input of the $p^{th}$ layer indexed by the coordinates $i,j$ as $x^{p}(i,j)$, then a network with $n$ layers will output $y(i,j)=x^{n}(i,j)$. Mathematically, we can define the global receptive field as the range of pixels in $x^{0}$ that affects each $y(i,j)$. Intuitively, the global receptive field is the size of the region that a CNN uses towards making a single prediction.

More convolution layers and bigger sized filters allow small increases in the global receptive field but cause an exponential increase in computation time and memory requirements. A common practice in classification CNNs is the use  of strided pooling to downsample feature maps withing the network, allowing for much wider global receptive fields \cite{park2016look}. Pooling operations have also been reported to provide translation invariance to CNN models \cite{park2016look}. However, the properties that make pooling useful in classification tasks are not desirable for stereo matching, so most stereo algorithms avoid this operation. The loss of detail from feature downsampling makes it harder to recognize very small differences, something crucial for pixel-level matching. We address this problem by using transpose convolution (deconvolution) operations. 

Deconvolution operations allow CNNs to learn filters capable of upsampling feature maps. The operation is especially useful in pixel-level applications, such as semantic segmentation or  generative networks. For example, for optical flow, where the matching search space is bidimensional,  the FlowNet \cite{fischer2015flownet} sequentially downsamples the features maps with pooling operations and uses a series of deconvolutions to obtain a dense prediction map. Unlike FlowNet, we argue that it is easier to match upsampled features than upsampling matching scores. Because of this we choose to implement deconvolution layers before computing any correlation metric. Just like represented in Figure \ref{fig:maingraph}, we implement the same amount of 2 strided $3\times{3}$ deconvolutions as the number of max poolings within the CNN. This creates a dense feature space that can be used for computation of a correlation score for every possible disparity level. 

\begin{figure*}
    \centering
    \includegraphics[width=\textwidth]{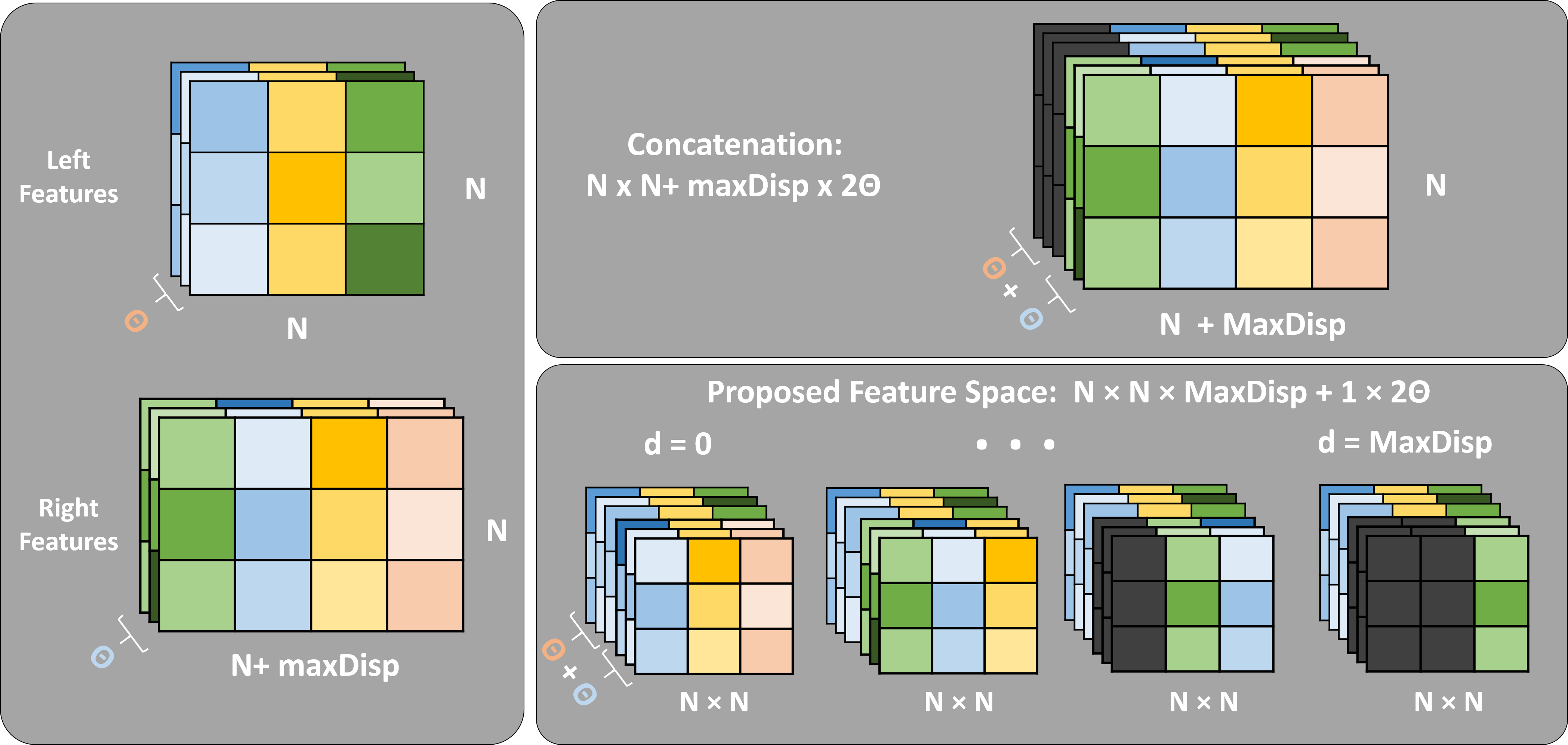}
    \caption{Comparison between standard feature concatenation and the proposed feature space. The left and right $\Theta$-dimensional features are computed by the siamese architecture. Similar color squares represent point correspondences between the stereo image pair. Differences in tone are just meant to represent small variations between both images. Black squares represent zero padding. }
    \label{fig:featurespace}
\end{figure*}

\subsection{Correlation layer}

Several stereo matching CNNs use the inner product as a correlation metric between features vectors extracted from the siamease branches \cite{luo2016efficient,zbontar2016stereo,park2016look}. The operation is computationally efficient, fast and differentiable, which allows backpropagation during training. In these cases, the CNN learns feature extractors that minimize the inner product between two corresponding points. While this provides a fast and effective way to compute correlation, it would be preferable to allow the network to learn a correlation that best fits the stereo data. Note that the inner product only measures one direction/component of similarity between vectors. Whereas the network could learn more complex relationships.

Recent methods choose to concatenate the output from the siamese network along the feature dimension and follow it with more convolution layers \cite{zbontar2016stereo,shaked2016improved,park2016look}. To a certain extent, this allows the CNN to learn how to correlate matching points, but the maximum disparity that the network is able to find is intrinsically related to the global receptive field of the layers stacked after the siamese portion of the CNN.

Lets consider the case where we want to find the disparity map for a left stereo image $I_{l}$ with $W\times{H}$ dimensions. Considering $D$, the maximum disparity possible between the stereo pair, correlation needs to be computed with all pixels within a $D+1$ range in the right stereo image $I_{r}$, just as described in Equation \ref{eq1}. By using a siamease network with a $\theta$ dimensional output its possible to extract two feature vectors with $W\times{H}\times{\theta}$ dimensions. To learn how to match pixels for $D+1$ possible disparities from the concatenated volume, the network needs to process $2\theta$ values in its third dimension and to account for a range of $D+1$ pixels in the input second dimension.  In other words, the correlation layers would need to start with $2\theta$ neurons, and their global receptive field would need to be equal or superior to $D+1$ in the image width dimension. Using the common approach where we stack $n$ layers of $w\times{w}$ convolution blocks the global receptive field of a network is equal to $n\times{(w-1)}+1$. In the KITTI dataset \cite{Geiger2012CVPR}, for example, where $D=256$, it would take at least 128 layers of $3\times{3}$ convolutions for a network to have a global receptive field wide enough to match 256 pixels apart without downsampling the feature space. This is not only challenging from a computational point of view but it greatly complicates the learning process. Beyond learning how to correlate features of matching points, the model would also need to correspond feature positions with the intended disparity. We propose a new correlation layer that greatly simplifies the learning problem, needing as little as two convolution layers to compute a disparity map for any size $D$.

Defining the  $\theta$-dimensional feature vectors computed from $I_{l}$ and $I_{r}$ as the $\psi_l$ and $\psi_r$, respectively, we construct a new feature space $\Psi$ as:
 \begin{equation}
\Psi(i,j)= \mid [ \mid \psi_l(i,j) \psi_r(i,j-d)\mid ] , \forall d \in  \mathbb{N}_0 \mid   0\leqslant d\leqslant D \mid \label{eq3}
\end{equation}
where $\mid . \mid$ represents a concatenation operation. Note that we are still concatenating vectors along the feature dimension, but we replicate the left features and pair them with right features of every possible disparity. The new feature space $\Psi$ has the dimensions $WH\times{D+1}\times{2\theta}$ where, for all  $(i,j)$ pixels, there is a paired $2\theta$-dimensional feature vector for all $D+1$ possible disparities. This simple transformation radically changes what kind of information convolution filters receive. 
Lets consider applying a single $1\times{1}$  convolution layer that outputs a single value from a $2\theta$ dimensional input to the new feature space $\Psi$. Note that a single value would be computed for $D+1$ disparities for all $(i,j)$ pixels, using only the corresponding right and left feature pairing as input. This way, the correlation layer only needs to learn how to correlate two concatenated $\theta$-dimensional vectors, independently of their original position, considerably simplifying the learning problem. This layer would output a $WH\times{D+1}\times{1}$ map that can be easily transformed to the intended disparity volume with a $W\times{H}\times{D+1}$ shape. Beyond this, in this feature space, filters of size $1\times{z}$ allow the network to learn a correlation metric that accounts for $z$ neighbor disparity pairs, creating the opportunity for a more robust disparity correlation.  Finally, because the filters learned during training always correlate $2\theta$-dimensional feature pairs, $\Psi$ can be rebuilt for a variable number of max disparities without needing to retrain the model.  

In our experimental results, we compare the performance of siamease architectures trained with inner product and with our correlation layer. We use the simplest architecture that allows non-linear logical operations \cite{rojas2013neural}. We use a single activated hidden layer with $2\theta$ neurons and $1\times{3}$ filters, and a single output neuron also with a $1\times{3}$ filter.

\begin{table*}[t]
\centering
\caption{Comparison of several error metrics in $\%$ of our three different siamese architectures trained with inner product (inner prod) and with our correlation architecture (learned)  on the KITTI 2012 validation set}
\label{table:kitti2012}
\resizebox{\textwidth}{!}{\begin{tabular}{cc|cccccc|c|}
\cline{3-9}
                                  &           & \multicolumn{2}{c|}{\textgreater 2 pixel}      & \multicolumn{2}{c|}{\textgreater 3 pixel}      & \multicolumn{2}{c|}{\textgreater 5 pixel}      & \multirow{2}{*}{Runtime (s)} \\ \cline{1-2}
\multicolumn{1}{|c}{Siamese CNN} & Correlation & \multicolumn{1}{c}{Non-Occ} & \multicolumn{1}{c|}{All} & \multicolumn{1}{c}{Non-Occ} & \multicolumn{1}{c|}{All} & \multicolumn{1}{c}{Non-Occ} & \multicolumn{1}{c|}{All} &                              \\ \hline\hline
\multicolumn{1}{|c}{\multirow{2}{*}{$S_4$}}  & inner prod      &         12.42                     &       14.18                   &                  11.38            &     13.16                   &                     9.98         &         11.76                 &        1.15                      \\ 
\multicolumn{1}{|c}{}                     & learned            &           11.27                   &  13.05                     &  10.39                            &       12.13               &   9.08                           &         10.82              &      5.25                        \\
\hline \hline
\multicolumn{1}{|c}{\multirow{2}{*}{$S_7$}}  & inner prod         &         7.57                     &     9.45                     &  6.72                            &   8.61                       &    5.64                          &     7.53                     &        1.15                      \\ 
\multicolumn{1}{|c}{}                     & learned            &            \textbf{ 6.65}                 &    \textbf{  8.23 }                      &          \textbf{  5.84   }               &    \textbf{  7.58}                        &                \textbf{    4.80  }        &       \textbf{ 6.48   }                 &      5.27                        \\ 
\hline\hline
\multicolumn{1}{|c}{\multirow{2}{*}{$S_9$}}                     & inner prod            & 7.47                             &     9.34                     &      6.50                        &        8.36                  &    5.31                      &         7.17                     &         1.16                    \\ 
\multicolumn{1}{|c}{}                     & learned            &        7.57                    &           10.29                &  6.59                       &     9.05                      &  5.34                  &             7.80              &               5.28               \\ \hline
\end{tabular}}

\end{table*}

\begin{figure*}[t]
    \centering
    \includegraphics[width=\textwidth]{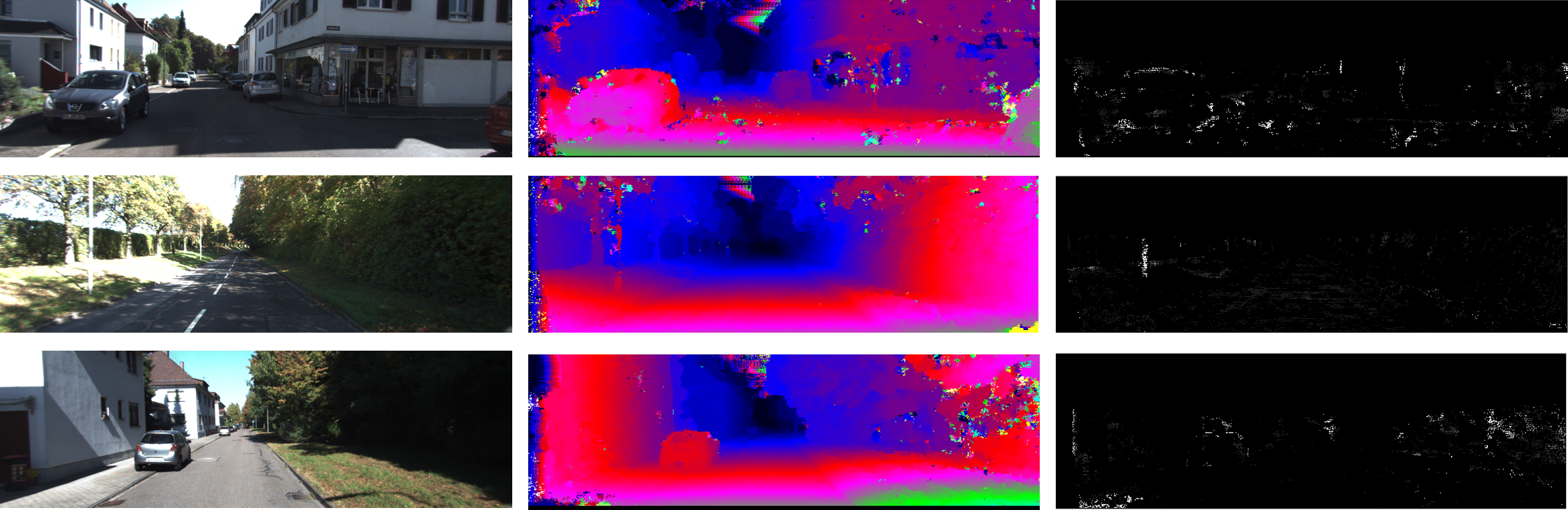}
    \caption{Examples of non-regularized disparities (middle) and errors (right) of KITTI 2012 validation images (left) computed with the $S_7$ architecture and learned correlation.}
    \label{fig:kitti2012}
\end{figure*}

\subsection{Training}

We train our models with stereo image pairs from the KITTI datatsets \cite{Geiger2012CVPR,Menze2015CVPR}, where the true displacement of a sparse number of pixels is known. We randomly extract small patches from the left stereo image and the same coordinate patch from the right image extended by the maximum disparity under consideration. This allow to diversely sample training batches while being memory efficient. We treat each disparity value as a mutually exclusive classification problem. The values outputted from the correlation step are used in a softmax loss. All parameters are trained with stochastic gradient descent and gradients are backpropagated using the standard Adam optimization \cite{DBLP:journals/corr/KingmaB14}.

\subsection{Testing}

During testing, memory constrains us to one-pass computations of disparity maps for high resolution images with big max displacements. Instead of processing subsections of the image individually, we follow the same procedure suggested by \cite{luo2016efficient}. First, we extract the feature representation for all pixels of the stereo image pair with the siamese architecture. Then in the correlation step, the same feature values can be reused for computation of disparity maps of multiple pixels. This results in significant increases in the inference speed.

\section{Experimental Evaluation}

We train and evaluate our models using both the KITTI 2012 \cite{Geiger2012CVPR} and KITTI 2015 \cite{Menze2015CVPR} datasets. Both are composed of rectified natural images captured by a stereo camera. KITTI 2012 consists only of static environments while moving objects are present in KITTI 2015. Just like most methods \cite{zbontar2016stereo,luo2016efficient,kendall2017end,shaked2016improved}, we use the sparse available labels from non-occluded pixels for training.

We evaluate our methodology by training three different siamese architectures: $S_4$, $S_7$ and $S_9$, with 4, 7 and 9 convolution layers and with 1, 2 and 3 max pooling layers, respectively. We also compare all models trained with inner product and with the proposed correlation architecture. 

All parameters are randomly initialized with a normalized Gaussian distribution and input images are normalized to have zero mean and unit standard deviation. Every CNN is trained for 75K iterations with a $1e^{-3}$ starting learning rate. Training is done with randomly extracted patches from left image with sizes $10\times{10}$ for $S_4$, $28\times{28}$ for $S_7$ and $56\times{56}$ for $S_9$. We use the biggest batch size that our system allowed for each model. For CNNs trained with inner product, this translates to batches of 128, 32 and 20  for $S_4$,$S_7$ and $S_9$, respectively, and batches of 128, 20 and 8 for the same models trained with out correlation architecture. All models were implemented in Tensorflow \cite{tensorflow2015-whitepaper} and ran on a NVIDIA Titax-X GPU.

\begin{table*}[t]
\centering
\caption{Comparison of several error metrics in $\%$ of our three different siamese architectures trained with inner product (inner prod) and with our correlation architecture (learned)  on the KITTI 2015 validation set}
\label{table:kitti2015}
\resizebox{\textwidth}{!}{\begin{tabular}{cc|cccccc|c|}
\cline{3-9}
                                  &           & \multicolumn{2}{c|}{\textgreater 2 pixel}      & \multicolumn{2}{c|}{\textgreater 3 pixel}      & \multicolumn{2}{c|}{\textgreater 5 pixel}      & \multirow{2}{*}{Runtime (s)} \\ \cline{1-2}
\multicolumn{1}{|c}{Siamese CNN} & Correlation & \multicolumn{1}{c}{Non-Occ} & \multicolumn{1}{c|}{All} & \multicolumn{1}{c}{Non-Occ} & \multicolumn{1}{c|}{All} & \multicolumn{1}{c}{Non-Occ} & \multicolumn{1}{c|}{All} &                              \\ \hline\hline
\multicolumn{1}{|c}{\multirow{2}{*}{$S_4$}}  & inner prod      &       11.19                       &  12.68                        &     10.01                      &           11.50                 &    8.57                          &                10.05         &      1.15                        \\ 
\multicolumn{1}{|c}{}                     & learned            &      8.26                        &             10.72             &   7.10                           &           9.71               &     6.82                         &       8.40                   &                        5.25      \\
\hline \hline
\multicolumn{1}{|c}{\multirow{2}{*}{$S_7$}}  & inner prod         &    7.80                          &   9.36                       &             6.81                 &      8.37                    &    5.75                          &      7.30                    &   1.15                            \\ 
\multicolumn{1}{|c}{}                     & learned            &   \textbf{6.79}                           &  \textbf{8.21 }                       &     \textbf{ 5.92 }                      &     \textbf{7.30 }                     &   \textbf{4.92   }                        &     \textbf{6.24    }                 &    5.27                         \\ 
\hline\hline
\multicolumn{1}{|c}{\multirow{2}{*}{$S_9$}}  & inner prod         &      6.89                        & 8.47                         &           6.02                   &    7.61                      &    5.18                          &      6.74                    &         1.16                     \\ 
\multicolumn{1}{|c}{}                     & learned            &        7.47                      &   8.96                     &    6.42                           &  7.88                         &   5.41                           &    6.82                      &          5.28                    \\ \hline
\end{tabular}}

\end{table*}

\begin{figure*}
    \centering
    \includegraphics[width=\textwidth]{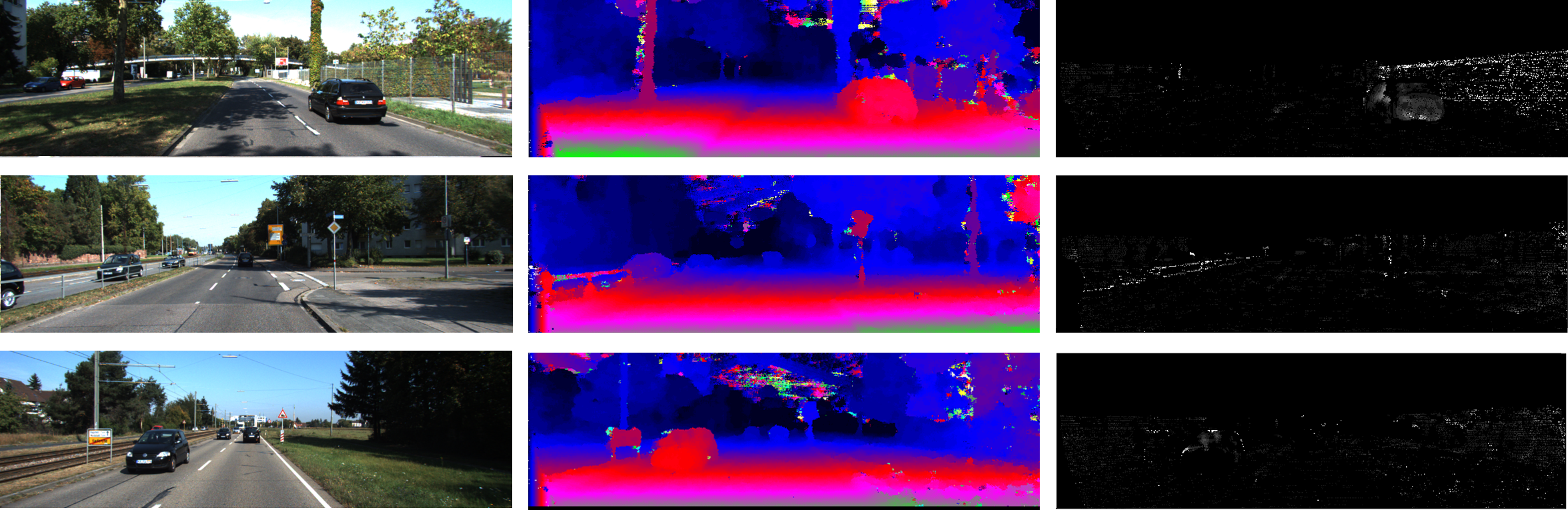}
    \caption{Examples of non-regularized disparities (middle) and errors (right) of KITTI 2015 validation images (left) computed with the $S_7$ architecture and learned correlation.}
    \label{fig:kitti2015}
\end{figure*}

\subsection{KITTI 2012}

The KITTI 2012 datasets consists of 194 image pairs for training and 195 for testing. Because no ground truth is given for the testing images, and multiple online submissions are not allowed, we evaluate our models by spliting the training data in a training and validation sets. As in the work developed by \cite{luo2016efficient}, we randomly use 160 image pairs for training and 34 for testing. Again, our main objective is to study and improve the siamease architecture that initializes most recent CNN stereo matching systems, so we do not implement an end-to-end system capable of competing with current state-of-the-art systems. The performance of our models in the validation set is shown in Table \ref{table:kitti2012}.

When we use the inner product for feature correlation, a direct comparison with the same depth architectures from \cite{luo2016efficient} allow us to verify the effect of pooling and deconvolution layers. All our models outperform the corresponding networks proposed by \cite{luo2016efficient}, which shows the benefit of our pooling/deconvolution approach. Despite the overall increase in performance, Table \ref{table:kitti2012} shows that there is a limit to the benefit of increasing the receptive field trough downsampling pooling layers. While  the 2-pixel is reduced substantially from  $S_4$ to $S_7$, the extra pooling layers in $S_9$ did not greatly decreased the matching error.

Table \ref{table:kitti2012} also shows that slightly better matching was achieved by learning correlations from the proposed feature space. While the overall increase in performance is small, the correlation layer substantially improves matching in edge regions when compared to the inner product counterpart. Matching improvements are present in $S_4$ and $S_7$ when the correlation layer is used, but a slightly worst performance is achieved in $S_9$. This indicates that the loss of detail from successive pooling might hinder the ability of the network to learn a good correlation function. The best results were achieved with $S_7$, where the receptive field is big enough for robust matching, but the lost of detail is not enough to stop the network from computing an effective correlation. Figure \ref{fig:kitti2012} shows that, even without spatial regularization, our architecture is able to smoothly match low detail regions while maintaining sharp edges in cars and trees.

\subsection{KITTI 2015}
KITTI 2015 has 200 image pairs for training and for testing. Again, just like \cite{luo2016efficient},  we randomly split the training set in 160 images for training and 40 for validation. This allows a better direct comparison with their method.

A similar analysis to the one  made for KITTI 2012 is valid for the KITTI 2015 results. Bigger receptive fields allow lower matching errors for features learned with the inner-product implementation. When learning a correlation, a compromise between a wider global receptive field with less loss of detail is found in the $S_7$ architecture. In Figure \ref{fig:kitti2015}, we continue to predict big smooth disparities in low texture regions, even without any post-processing. 
This shows that wider global receptive fields allow a much more effective correlation computation. Furthermore, even with the downsampling operation within the networks, features capable of representing small structures like traffic signs, fences and trees can be successfully extracted. Stacking further layers should easily allow spatial regularization to be learned without significant increase in computation cost, since the concatenation and reshaping operations of the feature space transformation are the bottleneck of the method.

\subsection{Comparisons with other methods}

\begin{table}
\centering
\caption{Comparison of the 2 pixel $\%$ error of different matching siamease architectures without post-processing on the 2012 and 2015 KITTI validation set}

\label{tab:compar}
\begin{tabular}{|c|cccc|}
\hline
\multirow{2}{*}{Method} & \multicolumn{2}{c|}{KITTI 2012} & \multicolumn{2}{c|}{KITTI 2015} \\ \cline{2-5} 
                        & Non-Occ        & All            & Non-Occ        & All            \\  \hline \hline
MC-CNN-acrt             & 15.02          & 16.92          & 15.20          & 16.83          \\
MC-CNN-fast             & 17.72          & 19.56          & 18.47          & 20.04          \\
Luo et al.              & 10.87          & 12.86          & 9.96           & 11.67          \\
$S_9$ + inner product     & 7.57 & 10.29  & 6.89  & 8.47  \\
$S_7$ + correlation     & \textbf{6.65}  & \textbf{8.23}  & \textbf{6.79}  & \textbf{8.21}  \\ \hline
\end{tabular}

\end{table}
As stated before, we do no propose a full stereo pipeline for stereo matching. Our main objective is to study and improve a crucial part of most of the current CNN stereo matching models: the siamease architecture. Because of this, we compare our work with other non-spatial regularized architectures. This results are presented in Table \ref{tab:compar}.

Table \ref{tab:compar} shows that when compared with other non regularized Siamese architectures, our wider models have a significantly lower 2-pixel error in both 2012 and 2015 KITTI datasets. Furthermore, the proposed space transformation allows $S_7$ to learn a shallow correlation layer which allows it to outperform all other siamese architectures. 

The results reported do not guarantee that replacing the siamease architectures of more complex models, such as the one proposed by \cite{kendall2017end}, will improve matching performance, but they show promising potential even without spatial regularization. If nothing else, our models, just like the ones proposed by \cite{luo2016efficient}, provide a simple, fast and easy to train approach, but much more accurate results.

\section{Conclusion}

Similar to so many areas in computing, deep learning has allowed us to move at an  incredible speed towards a robust solution for stereo matching. As computation power increases, there is a natural tendency to move to bigger and more complex CNN models. In this work we demonstrated that big improvements are still possible by small, problem-specific adaptations that simplify the learning problem. For future work, we plan to incorporate the recent approaches that take use context for regularization, allowing us to take full advantage of the proposed feature extractor.

\bibliographystyle{unsrt} 
\bibliography{egbib}

\end{document}